\documentclass[10pt, english, twocolumn]{article}
\usepackage[a4paper,
            bindingoffset=0.2in,
            left=0.5in,
            right=0.7in,
            top=1in,
            bottom=1in,
            footskip=.25in]{geometry}
\usepackage{graphicx} 
\usepackage{amsmath}
\usepackage{booktabs}
\usepackage{siunitx}
\usepackage{multirow}
\usepackage{amssymb}
\usepackage{pifont}
\usepackage[font=small,labelfont=bf,tableposition=top]{caption}
\usepackage{float}
\usepackage{stfloats}
\usepackage[colorlinks]{hyperref}
\newcommand{\cmark}{\ding{51}}%
\newcommand{\xmark}{\ding{55}}%
\usepackage[backend=biber]{biblatex}   
\graphicspath{{./images/}}

\title{ToonOut: Fine-tuned Background-Removal for Anime Characters}
\author{Matteo Muratori\textsuperscript{1}\textsuperscript{2}, Joël Seytre\textsuperscript{1}\\
\small{matteo.muratori9@studio.unibo.it, joel@kartoon.ai}\\
\small{\textsuperscript{1}Kartoon AI, \textsuperscript{2}University of Bologna}}

\date{
    \begin{figure}[H]
        \centering
        \vspace{-0.6cm} 
        \includegraphics[width=\textwidth]{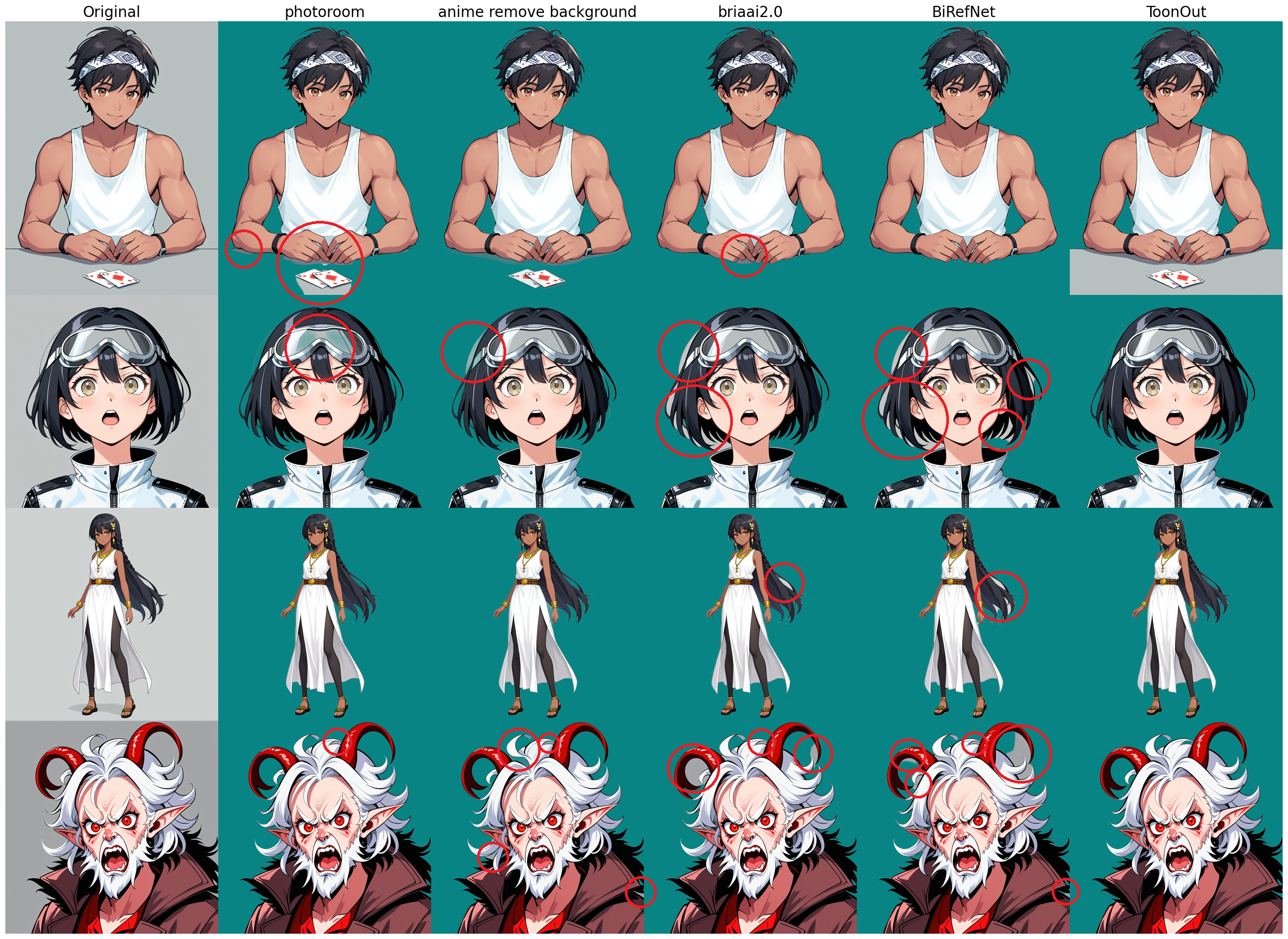}
        \onecolumn
        \vspace{-0.8cm} 
        \caption{\textbf{Closed source \cite{Photoroom} and open source models (\cite{AnimeRemoveBackground}, \cite{Briaai2.0}, \cite{BiRefNet}) are inadequate on our anime characters test set. We fine-tune BiRefNet \cite{BiRefNet} and the resulting model \textit{ToonOut} yields better outputs for our use-case.}}
        \label{fig:models_comparison}
        \twocolumn
        \vspace{-0.5cm} 
    \end{figure}
}

\begin{document}

\maketitle

\begin{center}
\section*{Abstract}
\end{center}

\textit{While state-of-the-art background removal models excel at realistic imagery, they frequently underperform in specialized domains—such as anime-style content, where complex features like hair and transparency present unique challenges. To address this limitation, we collected and annotated a custom dataset of 1,228 high-quality anime images of characters and objects, 
fine-tuned the open-sourced BiRefNet model on this dataset, resulting in marked improvements in background removal accuracy for anime-style images, going from 95.3\% to 99.5\% for our newly introduced Pixel Accuracy metric.
We are open-sourcing the code, the fine-tuned model weights, as well as the dataset at: \url{https://github.com/MatteoKartoon/BiRefNet}.}

\section{Introduction}
\begin{figure*}[t!]
    \centering
    \includegraphics[width=\linewidth]{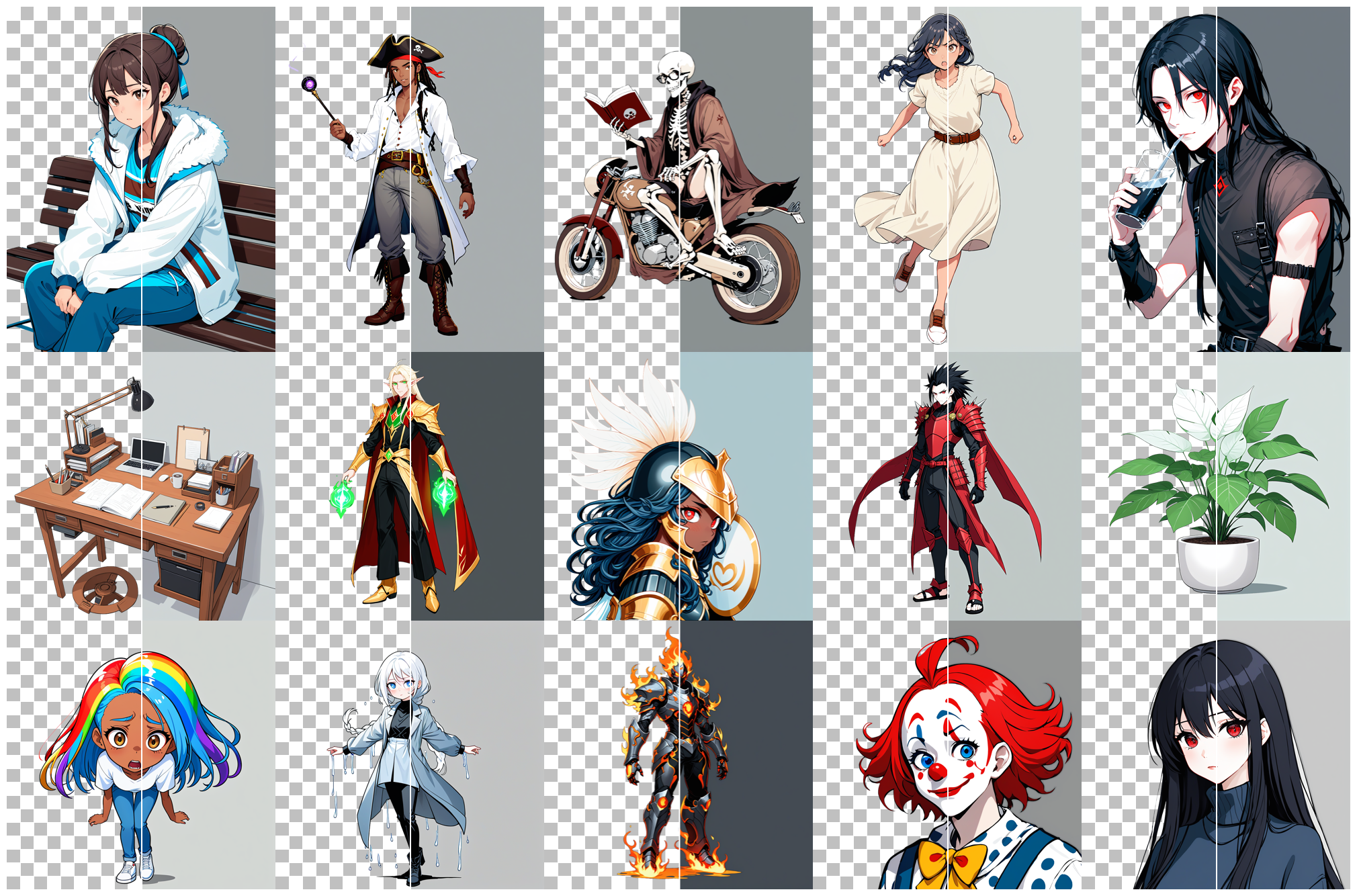}
    \caption{Examples of predictions made by \textit{ToonOut} on images contained in our test datasets.\\Our datasets cover a variety of characters in challenging poses, their interactions with items, and standalone objects.}
    \label{fig:model_predictions}
\end{figure*}

Recent advancements in machine learning have led to significant progress in various image processing tasks, including Dichotomous Image Segmentation (DIS) \cite{DIS}, which classifies image pixels into two categories (e.g., foreground and background).
DIS is commonly used for background removal, isolating the salient foreground object for downstream applications like image editing, story creation, or logo design.
State-of-the-art DIS models \cite{BiRefNet, Briaai2.0, Photoroom} demonstrate impressive capabilities, but their performance is degraded on anime-style images, such as those used at \textit{toongether} \cite{toongether} (see Figure~\ref{fig:models_comparison}). 
Even models specifically designed for anime \cite{AnimeBackgroundRemoval, AnimeRemoveBackground} often yield unsatisfactory results.\\

This challenge motivated our work to improve background removal for images of anime characters and items. 
We aimed to select a popular DIS model and enhance its capabilities on anime content. 
Our primary contributions are: (1) gathering a high-quality custom dataset of anime images depicting characters and items; 
(2) fine-tuning the popular BiRefNet model on this dataset, demonstrating a performance that matches the best closed-source models (see Table \ref{tab:overall_model_performances}); and (3) we introduce a new metric, \textit{Pixel Accuracy}, to evaluate fine-grained performance of DIS models.

\section{Dataset}

\subsection{Data sourcing}

\begin{figure*}[t!]
    \centering
    \includegraphics[width=\textwidth]{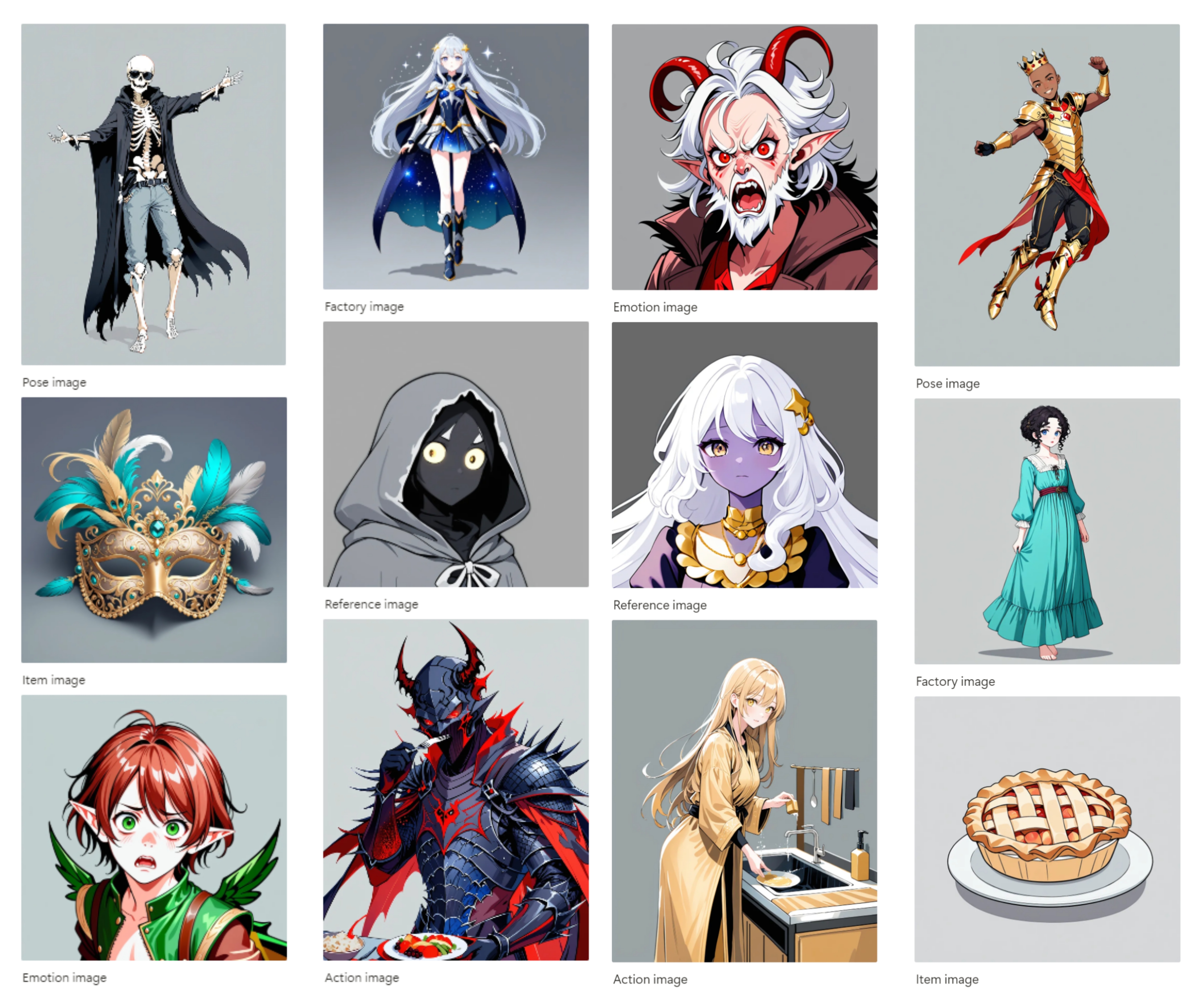}
    \caption{Example of images contained in the dataset, covering different characters, poses and items.}
    \label{fig:dataset_examples}
\end{figure*}

Our dataset consists of 1228 images broken down into \texttt{train / validation / test} with an \texttt{80\% / 10\% / 10\%} ratio (see Table \ref{tab:dataset_composition}). For each image, we provide both the original RGB image and the corresponding pixel-level ground truth mask. This consists of a grayscale image where black pixels represent background, white pixels represent foreground, and intermediate gray values indicate partially transparent pixels (useful in particular to provide a transparency gradient around character edges). \\

We designed our dataset to meet the following essential criteria:
\begin{itemize}
    \item \textit{Domain coverage:} The dataset should comprehensively represent the target domain, encompassing diverse anime-style content including both character portraits and object illustrations. To ensure broad generalization, we collected images featuring varied character designs, poses, viewing angles, and activities, grouped in various sub-datasets;
    \item \textit{Data quality:} All images are generations of good quality and high resolution (minimum 1024×1024 pixels) to preserve fine-grained details essential for accurate segmentation;
    \item \textit{Sample diversity:} To prevent overfitting and ensure robust evaluation metrics, the dataset presents high diversity. Each image represents a unique combination of character, pose, and context, so that we verify that the trained model doesn't overfit the \texttt{train} \& \texttt{validation} distribution.
\end{itemize}

We generated images using the anime-specialized checkpoint \textit{Yamer's Anime} \cite{yamers} of \textit{Stable Diffusion XL} \cite{SDXL, diffusion_models}. Generated images were rigorously filtered to ensure high quality: we discarded images with anatomical inconsistencies, unclear foreground-background boundaries, and artifacts that would result in visually unappealing segmentation masks.

Following the filtering phase, we evaluated the remaining images using the baseline model \textit{BiRefNet} to identify challenging cases where the baseline model performs poorly. We prioritized these difficult examples for dataset inclusion, as they represent the most valuable training samples for improving model performance. To maintain dataset balance, we also included up to 20\% of images where the baseline \textit{BiRefNet} model already performs well. \\

\begin{table*}[t!]
    \centering
    \resizebox{\textwidth}{!}{%
\begin{tabular}{lccccccccc}
    \toprule
    \multirow{2}{*}{\textbf{Model}} & \multirow{2}{*}{\textbf{Open code and weights}} & \multirow{2}{*}{\textbf{Open data}} & \multicolumn{7}{c}{\textbf{Performance over our test set (126 images)}} \\
    \cmidrule(lr){4-10}
    & & & \textbf{Pixel Accuracy} & \textbf{Mean Boundary IoU} & \textbf{Weighted F-measure} & \textbf{E-measure} & \textbf{S-measure} & \textbf{MAE} & \textbf{F-measure} \\
    \midrule
    Photoroom               & \xmark & \xmark & 99.2\% & 95.2\% & \textbf{99.3\%} & 99.2\% & 98.7\% & \textbf{0.04} & \textbf{99.2\%} \\
    Briaai2.0               & \cmark & \xmark & 97.8\% & 92.4\% & 98.8\% & 98.8\% & 97.9\% & 0.08 & 98.7\% \\
    BiRefNet                & \cmark & \cmark & 95.3\% & 88.5\% & 97.8\% & 97.9\% & 96.9\% & 0.15 & 98.0\% \\
    \textbf{ToonOut [ours]} & \cmark & \cmark & \textbf{99.5\%} & \textbf{95.6\%} & \textbf{99.4\%} & \textbf{99.4\%} & \textbf{98.9\%} & \textbf{0.03} & \textbf{99.2\%} \\
    \bottomrule
\end{tabular}%
}
    \caption{On our test sets, we manage to slightly outperform Photoroom's state-of-the-art remove-background model.}
    \label{tab:overall_model_performances}
\end{table*}

\subsection{Dataset Composition}
To ensure comprehensive domain coverage and training diversity, we structured our data collection around six distinct image categories, each targeting specific visual characteristics and segmentation challenges. The distribution and characteristics of each category are as follows (see Table \ref{tab:dataset_composition}):

\begin{itemize}
    \item \textit{Reference (neutral face portraits):} High-quality character portraits, with neutral emotion;
    
    \item \textit{Emotion (close-up portraits):} Character close-ups expressing various emotions (joy, anger, sadness, etc.). These images are a challenge for baseline models due to fine facial details and hair complexity;
    
    \item \textit{Pose (full-body characters in motion):} Full-body character representations in diverse poses (standing, jumping, running, etc.). Baseline model performance is generally better for these samples;
    
    \item \textit{Factory (full-body idle characters):} Full-body character representations in idle stance. Similarly to pose, the baseline performance is good;
    
    \item \textit{Action (characters with items):} Characters engaged in activities (cooking, gaming, working) that include objects. These samples represent the most challenging segmentation scenarios due to complex foreground-background interactions;
    
    \item \textit{Items:} Standalone object illustrations (vehicles, food items, tools, etc.). Baseline model performance varies significantly based on object complexity and boundary definition.
\end{itemize}

\section{Model: BiRefNet fine-tuning}
\subsection{BiRefNet}
For our fine-tuning approach, we selected the popular and high-performing \textit{Bilateral Reference Network (BiRefNet)}\cite{BiRefNet} as our base model. BiRefNet's DIS architecture incorporates several design principles well-suited to our segmentation objectives.
\textit{BiRefNet}'s bilateral reference mechanism employs dual supervision strategies: auxiliary gradient supervision \cite{Gradient} to enhance detail preservation in fine-grained regions, complemented by ground truth supervision, particularly useful in regions where foreground elements resemble the background in terms of color and texture. This dual supervision enables the model to maintain awareness of both local details and global context.
This aligns with our requirement that the model handle varying scales, from fine details of character hair to large-scale items and bodies.\\

BiRefNet’s architecture comprises two primary components: a \textit{Localization Module} that leverages global semantic information for object localization, and a \textit{Reconstruction Module} that reconstructs the segmentation using hierarchical image patches as the source reference \cite{Multi-view} and gradient maps as the target reference.

\begin{table*}[htbp]
    \centering
    \footnotesize
    \begin{tabular}{llrccc}
        \toprule
        \textbf{model}       & \textbf{dataset} & \textbf{\# images} & \textbf{Pixel Accuracy} & \textbf{Mean Boundary IoU} & \textbf{Weighted F-measure} \\
        \midrule
        Photoroom            &                  &                     & 98.9\%          & 90.9\%          & 99.3\%          \\
        BiRefNet            & reference        & 8                   & 96,6\%          & 82.7\%          & 98.5\%          \\
        \textbf{ToonOut [ours]} &              &                     & \textbf{99.8\%} & \textbf{95.5\%} & \textbf{99.7\%} \\
        \midrule
        Photoroom            &                  &                     & 99.7\%          & 95.0\%          & 99.6\%          \\
        BiRefNet            & emotion          & 26                  & 97.7\%          & 87.4\%          & 98.7\%          \\
        \textbf{ToonOut [ours]} &              &                     & \textbf{100.0\%}& \textbf{96.9\%} & \textbf{99.8\%} \\
        \midrule
        \textbf{Photoroom}   &                  &                     & \textbf{99.9\%} & \textbf{96.7\%} & \textbf{99.5\%} \\
        BiRefNet            & pose             & 25                  & 99.1\%          & 94.2\%          & 99.1\%          \\
        ToonOut [ours]       &                  &                     & \textbf{99.8\%} & 96.4\%          & \textbf{99.5\%} \\
        \midrule
        \textbf{Photoroom}   &                  &                     & \textbf{99.9\%} & \textbf{96.8\%} & \textbf{99.5\%} \\
        BiRefNet            & factory          & 41                  & 98.7\%          & 93.0\%          & 98.9\%          \\
        ToonOut [ours]       &                  &                     & \textbf{99.8\%} & 96.0\%          & \textbf{99.4\%} \\
        \midrule
        Photoroom            &                  &                     & 96.3\%          & 91.5\%          & 98.6\%          \\
        BiRefNet            & action           & 15                  & 76.8\%          & 69.4\%          & 91.2\%          \\
        \textbf{ToonOut [ours]} &              &                     & \textbf{99.0\%} & \textbf{93.1\%} & \textbf{99.3\%} \\
        \midrule
        \textbf{Photoroom}   &                  &                     & \textbf{98.3\%} & \textbf{94.3\%} & \textbf{98.6\%} \\
        BiRefNet            & items            & 11                  & 92.2\%          & 92.2\%          & 97.2\%          \\
        ToonOut [ours]       &                  &                     & 96.6\%          & 92.5\%          & 97.8\%          \\
        \midrule
        Photoroom            &                  &                     & 99.2\%          & 95.2\%          & \textbf{99.3\%} \\
        BiRefNet            & overall          & 126                 & 95.3\%          & 88.5\%          & 97.8\%          \\
        \textbf{ToonOut [ours]} &              &                     & \textbf{99.5\%} & \textbf{95.6\%} & \textbf{99.4\%} \\
        \bottomrule
    \end{tabular}
    \caption{Performance on the sub-datasets released. \textit{Items} remain challenging as they are a small part of the overall dataset.}
\label{tab:per_dataset_model_performances}
\end{table*}

\subsection{Fine-tuning experiment}
To fine-tune the \textit{BiRefNet} model we take as a starting point a checkpoint saved after 244 training epochs, using those weights to resume the training process.
The hyperparameters were selected based on extensive empirical experimentation. The final settings we used are as follows:
\begin{itemize}
    \item \textit{loss weights:} the loss function is a weighted sum between \textit{SSIM loss, MAE loss, and IoU loss}. A detailed explanation about these losses is provided at \cite{BiRefNet} \cite{MAE} \cite{Losses}. 
    \begin{equation}
        L=\lambda_{1}L_{SSIM}+\lambda_{2}L_{MAE}+\lambda_{3}L_{IoU}
    \end{equation}
    where
    $\lambda_{1}=10, \ \ \lambda_{2}=90, \ \ \lambda_{3}=0.25 \;$
    \item additionally, we use a binary cross-entropy loss to supervise the gradients;
    \item \textit{training:} we train on 2 GeForce RTX 4090s for $46$ epochs, with a batch size of 2. The learning rate started from $1e-5$, and was reduced by half after $20$ and $40$ epochs. We clip the gradients at $100$.
\end{itemize}

\section{Results}
\subsection{Metrics overview}
First, we select appropriate metrics to assess the performance of background removal in anime character images.
Empirically, two primary categories of model prediction errors can be observed:
\begin{itemize}
    \item \textbf{Coarse-grained errors}: These occur when large regions are incorrectly classified. Common examples include the area between a character’s ponytail and head, between limbs and torso, folds in clothing, or the character’s shadow.
    \item \textbf{Fine-grained errors}: Small-scale inaccuracies, such as hair strands, fingers, detailed object contours.
\end{itemize}

Accordingly, evaluation measures for dichotomous image segmentation (DIS) fall into two categories:
\begin{itemize}
    \item \textbf{Coarse-grained metrics} judge errors at the pixel level, so small boundary mistakes barely change the score. We considered: \textit{S-measure} \cite{S-measure}, \textit{E-measure} \cite{E-measure}, \textit{F-measure}, \textit{Mean Absolute Error (MAE)} and \textit{Mean Squared Error (MSE)};
    \item \textbf{Fine-grained metrics} focus on pixels near object edges, rewarding crisp outlines. We looked at: \textit{Boundary Intersection Over Union (BIoU)} \cite{BIoU}, \textit{Human Correction Effort (HCE)} \cite{DIS}, \textit{Weighted F-measure (WF)} \cite{WF-measure}, and \textit{Mean Boundary Accuracy (MBA)} \cite{mBa}.
\end{itemize}

To select the most suitable evaluation metric, we applied the nine published measures—already implemented in the BiRefNet code—to a validation set. For each image, we first established a human-judged ranking of model outputs, then observed how often each metric aligned with that preference. The Boundary IoU (BIoU) demonstrated the strongest concordance with our visual assessments of boundary quality, whereas none of the region-oriented metrics aligned with our expectations.

\subsection{Pixel Accuracy}
That is why we introduce \textit{Pixel Accuracy (PA)}, meant to capture the percentage of correctly labeled pixels, as specified in equation \ref{eq:pixel_acc}.

A pixel is counted as correct when the absolute difference $\delta$ between prediction and ground-truth alpha values is $\delta \leq 10$, which is visually indistinguishable. 
To avoid penalizing harmless one-pixel-wide edge artifacts around the character and object boundaries, we erode the error mask once \cite{erode}. The \textit{PA} score is then computed as the relative fraction of correct pixels.

\begin{equation}
    PA = \left( 1-\frac{\text{Number of Incorrect Pixels}}{\text{Total Number of Foreground Pixels}} \right) ^2
    \label{eq:pixel_acc}
\end{equation}

where "foreground pixels" denote the pixels with $\alpha > 128$.

\subsection{Evaluating model performance}

After selecting the optimal checkpoint based on the validation set, we evaluate the performance of our \textit{ToonOut} model on the introduced test sets.

This evaluation follows three key metrics: \textit{Pixel Accuracy (PA)}, \textit{Boundary Intersection over Union (BIoU)}, and \textit{Weighted F-measure (WF)}. We benchmarked \textit{ToonOut} against two prominent models: the closed-source high-performing \textit{Photoroom} \cite{Photoroom} and two open-source baselines, \textit{Briaai2.0} \cite{Briaai2.0} and \textit{BiRefNet} \cite{BiRefNet}.

The aggregated results, presented in Table \ref{tab:overall_model_performances}, indicate a clear trend: \textit{ToonOut} bridges the gap between the open-source models and \textit{Photoroom}, overall reducing the \textit{PA error rate} from ~2.2\% / ~4.7\% to ~0.5\%, and the \textit{BIoU error rate} from ~7.6\% / ~11.5\% to ~4\%.

Looking across individual datasets (Table \ref{tab:per_dataset_model_performances}), the performance is slightly worse for the \textit{items} category, where the images are few (only 107) and varied. Overall, we can observe the high performance of \textit{ToonOut}'s capabilities in anime character background removal, particularly in the challenging datasets \textit{action} and \textit{emotion}.

\begin{table}[htbp]
    \centering
    \begin{tabular}{@{} l c c c c r @{}}
        \toprule
        \multirow{2}{*}{\textbf{Dataset}} & \multicolumn{4}{c}{\textbf{number of images}} & \multirow{2}{*}{\textbf{total \%}}\\
        \cmidrule(lr){2-5}
        & \textbf{train} & \textbf{validation} & \textbf{test} & \textbf{total} & \\
        \midrule
        Reference          & 58            & 7                  & 8             & 73            & 5.9\% \\
        Emotion            & 201           & 25                 & 26            & 252           & 20.5\% \\
        Pose               & 199           & 25                 & 25            & 249           & 20.3\% \\
        Factory            & 324           & 41                 & 41            & 406           & 33.1\% \\
        Action             & 112           & 14                 & 15            & 141           & 11.4\% \\
        Items              & 85            & 11                 & 11            & 107           & 8.7\% \\
        \midrule
        \textbf{Total}     & \textbf{979}  & \textbf{123}       & \textbf{126}  & \textbf{1228} & \textbf{100\%} \\
        \bottomrule
    \end{tabular}
    \caption{Composition of our datasets.}
    \label{tab:dataset_composition}
\end{table}

\section{Conclusion}

In this paper, we addressed the challenging task of accurate background removal in anime character images, a critical preprocessing step for numerous creative downstream applications. We gathered, annotated and released a novel high-quality dataset, and fine-tuned BiRefNet on this specific task.

Evaluated on the novel evaluation metric \textit{Pixel Accuracy} as well as established metrics,
ToonOut outperforms both open-source baselines and the best closed-source model on anime character background-removal.

More broadly, our results highlight a key insight: with minimal effort, domain-specific fine-tuning can outperform general-purpose segmentation models in specialized contexts.

\section*{Acknowledgements}
Thanks to Gabriel Aubourg for going through the manual labor of providing high-quality annotations. Thanks to Cédric Roux, Samir Nasser Eddine, and Giovanni Paolini for helping this project happen.

\printbibliography

\end{document}